\def\year{2022}\relax
\documentclass[letterpaper]{article} 
\usepackage{aaai22}  
\usepackage{times}  
\usepackage{helvet}  
\usepackage{courier}  
\usepackage[hyphens]{url}  
\usepackage{graphicx} 
\usepackage{natbib}  
\usepackage{caption} 
\DeclareCaptionStyle{ruled}{labelfont=normalfont,labelsep=colon,strut=off} 
\usepackage{algorithm}
\usepackage{algorithmic}

\usepackage{xspace}
\usepackage{booktabs}
\usepackage{amsmath}
\usepackage{amssymb}

\newcommand\bert{\textsc{BERT}\xspace}
\newcommand\roberta{\textsc{RoBERTa}\xspace}

\newcommand\fullselect{\textsc{Full-Select}\xspace}
\newcommand\simselect{\textsc{Sim-Select}\xspace}
\newcommand\divselect{\textsc{Div-Select}\xspace}

\newcommand\news{\textsc{News}\xspace}
\newcommand\abstracts{\textsc{Abstracts}\xspace}
\newcommand\reports{\textsc{Reports}\xspace}

\newcommand\climatebert{\textsc{ClimateBert}\xspace}

\newcommand\climatefever{\textsc{climate-fever}\xspace}

\newcommand\fever{\textsc{fever}\xspace}
\newcommand\corpus{\textsc{Corp}\xspace}

\usepackage{newfloat}
\usepackage{listings}
\floatstyle{ruled}
\newfloat{listing}{tb}{lst}{}
\floatname{listing}{Listing}
\title{\climatebert: A Pretrained Language Model for Climate-Related Text}
\author{
    Nicolas Webersinke,\textsuperscript{\rm 1}
    Mathias Kraus,\textsuperscript{\rm 1}
    Julia Anna Bingler,\textsuperscript{\rm 2}
    Markus Leippold\textsuperscript{\rm 3}
}
\affiliations{
    \textsuperscript{\rm 1}FAU Erlangen-Nuremberg, Germany\\
    \textsuperscript{\rm 2}ETH Zurich, Switzerland\\
    \textsuperscript{\rm 3}University of Zurich, Switzerland\\
    nicolas.webersinke@fau.de, mathias.kraus@fau.de, binglerj@ethz.ch, markus.leippold@bf.uzh.ch

}

\begin{document}

\maketitle

\begin{abstract}
Over the recent years, large pretrained language models (LM) have revolutionized the field of natural language processing (NLP). However, while pretraining on general language has been shown to work very well for common language, it has been observed that niche language poses problems. In particular, climate-related texts include specific language that common LMs can not represent accurately. We argue that this shortcoming of today's LMs limits the applicability of modern NLP to the broad field of text processing of climate-related texts. As a remedy, we propose \climatebert, a transformer-based language model that is further pretrained on over 2 million paragraphs of climate-related texts, crawled from various sources such as common news, research articles, and climate reporting of companies. We find that \climatebert leads to a 48\% improvement on a masked language model objective which, in turn, leads to lowering error rates by 3.57\% to 35.71\% for various climate-related downstream tasks like text classification, sentiment analysis, and fact-checking.
\end{abstract}

\section{Introduction}
Researchers working on climate change-related topics increasingly use natural language processing~(NLP) to automatically extract relevant information from textual data. Examples include the sentiment or specificity of language used by companies when discussing climate risks and measuring corporate climate change exposure, which increases transparency to help the public know where we stand on climate change \citep[e.g.,][]{callaghan2021machine, bingler2022initiatives}. Many studies in this domain apply traditional NLP methods, such as dictionaries, bag-of-words approaches or simple extensions thereof \citep[e.g.,][]{gruning2011artificial, sautner2020firm}. However, such analyses face considerable limitations, since climate-related wording could vary substantially by source \citep{Kim2018}. Deep learning techniques that promise higher accuracy are gradually replacing these approaches \citep[e.g.,][]{kolbel2020ask, luccioni2020analyzing, bingler2022cheap, callaghan2021machine, chillrud2021evidence, friederich2021automated}. Indeed, it has been shown in related domains that deep learning in NLP allows for impressive results, outperforming traditional methods by large margins \citep{varini2020climatext}.

These deep learning-based approaches make use of language models (LMs), which are trained on large amounts of textual and unlabelled data. This training on unlabelled data is called \emph{pretraining} and leads to the model learning representations of words and patterns of common language. One of the most prominent language models is called \bert (Bidirectional Encoder Representations from Transformers) \citep{devlin2018bert} with its successors \roberta \citep{liu2019roberta}, Transformer-XL \citep{dai2019transformer} and ELECTRA \citep{clark2020electra}. These models have been trained on huge amounts of text which was crawled from an unprecedented amount of online resources.

After the pretraining phase, most LMs are trained on additional tasks, the \emph{downstream task}. For the downstream tasks, the LM builds on and benefits from the word representations and language patterns learned in the pretraining phase. The pre-training benefit is especially large on downstream tasks for which the collection of samples is difficult and, thus, the resulting training datasets are small (hundreds or few thousands of samples). Furthermore, it has been shown that a model that was pretrained on the downstream task-specific text exhibits better performance, compared to a model that has been pretrained solely on general text \citep{araci2019finbert, lee2020biobert}.

Hence, a straightforward extension to the standard combination of pretraining is the so-called domain-adaptive pretraining \citep{gururangan2020don}. This approach has recently been studied for various tasks and basically comes in the form of pretraining multiple times --- in particular pretraining in the language domain of the downstream task, i.e., 
\begin{align*}
    & \text{pretraining (general domain)} \\
    + \, & \textbf{domain-adaptive} \\
     \, & \textbf{pretraining (downstream domain)} \\
    + \, & \text{training (downstream task)}.
\end{align*}

To date, regardless of the increase in using NLP for climate change related research, a model with climate domain-adaptive pretraining has not been publicly available, yet. Research so far rather relied on models pretrained on general language, and fine-tuned on the downstream task. To fill this gap, our contribution is threefold. First, we introduce \climatebert, a state-of-the-art language model that is specifically pretrained on climate-related text corpora of various sources, namely news, corporate disclosures, and scientific articles. This language model is designed to support researchers of various disciplines in obtaining better performing NLP models for a manifold of downstream tasks in the climate change domain. Second, to illustrate the strength of \climatebert, we highlight the performance improvements using \climatebert on three standard climate-related NLP downstream tasks. Third, to further promote research at the intersection of climate change and NLP,
we make the training code and weights of all language models publicly available at GitHub and Hugging Face.\footnote{\url{www.github.com/climatebert/language-model}}\footnote{\url{www.huggingface.co/climatebert}}

\begin{figure*}
	\centering
	\includegraphics[width=.9\textwidth]{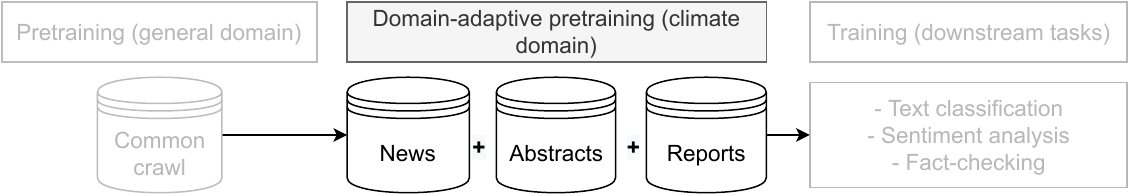}
	\caption{Sequence of training phases. Our main contribution is the continued pretraining of language models on the climate domain. In addition, we evaluate the obtained climate domain-specific language models on various downstream tasks.}
	\label{fig:framework}
\end{figure*}

\section{Background}
As illustrated in Figure \ref{fig:framework}, our LM training approach for \climatebert comprises all three phases --- using an LM pretrained on a general domain, the domain-adaptive pretraining on the climate domain, and the training phase on climate-related downstream tasks. 

\subsection{Pretraining on General Domain}

As of 2018, pretraining became the quasi-standard for learning NLP models. First, a neural language model, often with millions of parameters, is trained on large unlabeled corpora in a semi-supervised fashion. By learning on multiple levels which words/word-sequences/sentences appear in the same context, an LM can represent a semantically similar text by similar vectors. Typical objectives for training LMs are the prediction of masked words or the prediction of a label indicating whether two sentences occurred consecutively in the corpora \citep{devlin2018bert}. 

In the earlier NLP pretraining days, LMs traditionally used regular or convolutional neural networks \citep{collobert2008unified}, or later Long-Short-Term-Memory (LSTM) networks to process text \citep{howard2018universal}. Todays LMs mostly build on transformer models \citep[e.g.,][]{devlin2018bert, dai2019transformer, liu2019roberta}. One of the latter is named \roberta \citep{liu2019roberta} which was trained on 160GB of various English-language corpora - data from BOOKCORPUS \citep{zhu2015aligning}, WIKIPEDIA, a portion of the CCNEWS dataset \citep{nagel2016ccnews}, OPENWEBTEXT corpus of web content extracted from URLs shared on Reddit \citep{Gokaslan2019OpenWeb}, and a subset of CommonCrawl that is said to resemble the story-like style of WINOGRAD schemas \citep{Trinh2018ASM}. While these sources are valuable to build a model working on general language, it has been shown that domain-specific, niche language (such as climate-related text) poses a problem to current state-of-the-art language models \citep{araci2019finbert}.

\subsection{Domain-Specific Pretraining}
As a remedy to inferior performance of general language models when applied to niche topics, multiple language models have been proposed which build on the pretrained models but continue pretraining on their respective domains. FinBERT, LegalBert, MedBert are just a few language models that have been further pretrained on the finance, legal, or medical domain \citep{araci2019finbert, chalkidis2020legal, rasmy2021med}. In general, this domain-adaptive pretraining yields more accurate models on downstream tasks \citep{gururangan2020don}. 

Domain-specific pretraining requires a decision about which samples to include in the training process. It is still an open debate which sample strategy improves performance best. Various strategies can be applied to extract the text samples on which the LM is further pretrained. For example, while traditional pretraining uses all samples from the pretraining corpus, similar sample selection (\simselect) uses only a subset of the corpus, in which the samples are similar to the samples in the downstream task \citep{ruder2017learning}. In contrast, diverse sample selection (\divselect) uses a subset of the corpus, which includes dissimilar samples compared to the downstream dataset \citep{ruder2017learning}. Previous research has investigated the benefit of these approaches, yet no final conclusion about the efficiency has been obtained. Consequently, we compare these approaches in our experiments.

\subsection{NLP on Climate-Related Text}
In the past, climate-related textual analysis often used pre-defined dictionaries of presumably relevant words and then simply searched for these words within the documents. For example, \citet{cody2015sentiment} use such an approach for climate-related tweets. Similarly, \citet{sautner2020firm} use a keyword-based approach to capture firm-level climate change exposure. However, these methods do not account for context. The lack of context is a significant drawback, given the ambiguity of many climate-related words such as "environment," "sustainable," or "climate" itself \cite{varini2020climatext}.

Only recently, \bert has been used for NLP in climate-related text. The transformers-based \bert models are capable of accounting for the context of words and have outperformed traditional approaches by large margins across various climate-related datasets \citep{kolbel2020ask, luccioni2020analyzing, varini2020climatext, bingler2022cheap, callaghan2021machine, chillrud2021evidence, friederich2021automated, stammbach2022dataset}. However, this research has also shown that extracting climate-related information from textual sources is a challenge, as climate change is a complex, fast-moving, and often ambiguous topic with scarce resources for popular text-based AI tasks. 

While context-based algorithms like \bert can detect a variety of complex and implicit topic patterns in addition to many trivial cases, there remains great potential for improvement in several directions. To our knowledge, none of the above cited work has examined the effects of domain-adaptive pretraining on their specific downstream tasks. Therefore, we investigate whether domain-adaptive pretraining will improve performance for climate change-related downstream tasks such as text classification, sentiment analysis, and fact-checking.

\section{\climatebert}
In the following, we describe our approach to train \climatebert. We first list the underlying data sources before describing our sample selection techniques and, finally, the vocabulary augmentation we used for training the language model.

\subsection{Text Corpus}
Our goal was to collect a large corpus of text, \corpus, that included general and domain-specific climate-related language. We decided to include the following three sources: news articles, research abstracts, and corporate climate reports. We decided not to include full research articles because this language is likely too specific and does not represent general climate language. We also did not include Twitter data, as we assume that these texts are too noisy. In total, we collected 2,046,523 paragraphs of climate-related text (see Table \ref{tbl:datasets}).

The \news dataset is mainly retrieved from Refinitiv Workspace and includes 135,391 articles tagged with climate change topics such as climate politics, climate actions, and floods and droughts. In addition, we crawled climate-related news articles from the web.

The \abstracts dataset includes abstracts of climate-related research articles crawled from the Web of Science, primarily published between 2000 and 2019.

The \reports dataset comprises corporate climate and sustainability reports of more than 600 companies from the years 2015-2020  retrieved from Refinitiv Workspace and the respective company websites.

Given the nature of the datasets, we find a large heterogeneity between the paragraphs in terms of number of words. Unsurprisingly, on average, the paragraphs with the least words come from the \news and the \reports datasets. In contrast, \abstracts includes paragraphs with the most words. Table \ref{tbl:datasets} lists these descriptives.

To estimate the benefit from domain-adaptive pretraining, we compare the similarity of our text corpus with the one used for pretraining \roberta. Following \citet{gururangan2020don}, we consider the vocabulary overlap between both corpora. The resulting overlap of 57.05\% highlights the dissimilarity between the two domains and the need to add specific vocabularies. Therefore, we expect to see considerable performance improvements of domain-adaptive pretraining.

\begin{table}
\centering
\resizebox{\columnwidth}{!}{
\begin{tabular}{l cccc}
		\toprule
		    {\textbf{Dataset}} &
		    {\textbf{Num. of}} & \multicolumn{3}{c}{\textbf{Avg. num. of words}} \\
		    & {\textbf{paragraphs}} & {\textbf{Q1}} & {\textbf{Mean}} & {\textbf{Q3}} \\
		\midrule
        {News} & 1,025,412 & 34 & 56 & 65 \\
        {Abstracts} & 530,819 & 165 & 218 & 260 \\
        {Reports} & 490,292 & 34 & 65 & 79 \\ 
        \midrule
        {Total} & 2,046,523 & 36 & 107 & 168 \\
        \bottomrule
\end{tabular}}
\caption{Corpus \corpus used for pretraining \climatebert. Q1 and Q3 stand for the 0.25 and 0.75 quantiles, respectively.}
\label{tbl:datasets}
\end{table}

\subsection{Sample Selection}
Prior work has shown that specific selections of the samples used for pretraining can foster the performance of the LM. In particular, incorporating information from the downstream task by selecting similar or diverse samples has been shown to yield favorable results compared to using all samples from the dataset. We follow both approaches and select samples that are similar or diverse to climate-text using our text classification task (see \ref{sec:text_classification}). We experiment with three different strategies from \citet{ruder2017learning} for the selection of samples from our corpus:
\begin{itemize}
    \item In the most traditional sample selection strategy, \fullselect, we use all paragraphs from \corpus to train ${\climatebert}_F$.
    \item In \simselect, we select the 70\% of samples from \corpus, which are most similar to the samples of our text classification task. We use a Euclidean similarity metric for this sample selection strategy. We call this LM ${\climatebert}_S$.
    \item In \divselect, we select the 70\% of samples from \corpus, which are most diverse compared to the samples from our text classification task. We use the sum between the type-token-ratio and the Shannon-entropy for measuring diversity \citep{ruder2017learning}. This LM is named ${\climatebert}_D$.
    \item In \divselect + \simselect, we use the same diversity and similarity metrics as before. We then compute a composite score by summing over their scaled values. We keep the 70\% of the samples with the highest composite score to train ${\climatebert}_{D + S}$.   
\end{itemize}

\subsection{Vocabulary Augmentation}
We extend the existing vocabulary of the original model to include domain-specific terminology. This allows \climatebert to explicitly learn representations of terminology that frequently occur in a climate-related text but not in the general domain. In particular, we add the 235 most common tokens as new tokens to the tokenizer, thereby extending the size of the vocabulary for our basis language model (Distil\roberta) from 50,265 to 50,500. See Appendix \ref{sec:added_tokens} for a list of all added tokens. We also experimented with language models that do not use vocabulary augmentation or add more tokens. However, overall we find improvements using this technique and, thus, apply it to all language models which we pretrain on the climate domain.

\subsection{Model Selection}
For all our experiments, we use Distil\roberta, a distilled version of \roberta from Huggingface,\footnote{\url{www.huggingface.co/distilroberta-base}} as our starting point for training \citep{Sanh2019DistilBERTAD}. All our language models are trained with a masked language modeling objective (i.e., cross-entropy loss on predicting randomly masked tokens). We report all hyperparameters in Table \ref{tbl:hyperparamters}. The large batch size of 2016 for training the LM is achieved using gradient accumulation.

\subsection{Training on Downstream Task}
After pretraining Distil\roberta on \corpus, we follow standard practice \citep{devlin2018bert} and pass the final layer [CLS] token representation to a task-specific feedforward layer for prediction. We report all hyperparameters of this feedforward layer in Table \ref{tbl:hyperparamters}.

\begin{table}
\centering
\resizebox{\columnwidth}{!}{
\begin{tabular}{lcc}
        \toprule
		    & {\textbf{Downstream domain-}} & {\textbf{Downstream tasks}} \\
		    & {\textbf{adaptive pretraining}}& {\textbf{training}} \\
		\midrule
		\toprule
		    {\textbf{Hyperparameter}} & \multicolumn{2}{c}{Value} \\
		\midrule
        Batch size & 2016 & 32 \\
        Learning rate & 5e-4 & 5e-5 \\
        Number of epochs & 12 & 1000 \\
        Patience & --- & 4 \\
        Class weight & --- & Balanced \\
        Feedforward nonlinearity & --- & tanh \\
        Feedforward layer & --- & 1 \\
        Output neurons & --- & Task dependent \\
        Optimizer & \multicolumn{2}{c}{Adam} \\
        Adam epsilon & \multicolumn{2}{c}{1e-6} \\
        Adam beta weights & \multicolumn{2}{c}{(0.9, 0.999)} \\
        Learning rate scheduler & \multicolumn{2}{c}{Warmup linear} \\
        Weight decay & \multicolumn{2}{c}{0.01} \\
        \bottomrule
\end{tabular}}
\caption{Hyperparameters used for the downstream domain-adaptive pretraining and the downstream tasks training of \climatebert.}
\label{tbl:hyperparamters}
\end{table}

\section{Performance Analysis of Language Model}
Table \ref{tbl:lm_results} lists the results after pretraining Distil\roberta on \corpus with various sample selection strategies. For evaluation, we split \corpus randomly into 80\% training data and 20\% validation data. The reported loss is the cross-entropy loss on predicting randomly masked tokens from the validation data. We find that ${\climatebert}_F$ leads to the lowest validation loss. This performance is followed by the other \climatebert LMs, which all show similar results. Overall, we find that our domain-adaptive pretraining decreases the cross-entropy loss by 46--48\% compared to the basis Distil\roberta, cutting the loss almost in half. 

\begin{table}[h]
\centering
\begin{tabular}{l c}
		\toprule
		    {\textbf{Model}} &
		    {\textbf{Val. loss}} \\
		\midrule
        {Distil\roberta} & 2.238 \\[3pt]
        ${\climatebert}_F$ & 1.157 \\
        ${\climatebert}_S$ & 1.205 \\
        ${\climatebert}_D$ & 1.204 \\
        ${\climatebert}_{D+S}$ & 1.203 \\
        \bottomrule
\end{tabular}
\caption{Loss of our language models on a validation set from our text corpus \corpus.}
\label{tbl:lm_results}
\end{table}

\section{Performance Analysis for Climate-Related Downstream Tasks}

For our experiments, we used the following downstream tasks: text classification, sentiment analysis, and fact-checking. Table \ref{tbl:descriptives} lists basic statistics about the downstream tasks. We repeated the training and evaluation phase 60 times for each experiment, each time with a random 90\% set of samples for training and the remaining 10\% set for validation.

\begin{table}[h]
\centering
\resizebox{\columnwidth}{!}{
\begin{tabular}{lccc}
		\toprule
		    {\textbf{Downstream}} &
		    {\textbf{Num. of}} & {\textbf{Labels}} & {\textbf{Label}} \\
		    {\textbf{task}} & {\textbf{samples}} & & {\textbf{distribution}} \\
		\midrule
        {Text classification} & 1220 & climate-related: yes/no & 1000/220 \\
        {Sentiment analysis} & 1000 & opportunity/neutral/risk & 250/408/342 \\
        {Fact-checking} & 2745 & claim: support/refute & 1943/802 \\ 
        \bottomrule
\end{tabular}}
\caption{Overview of our downstream tasks used for evaluating \climatebert.}
\label{tbl:descriptives}
\end{table}

\subsection{Text Classification}
\label{sec:text_classification}
For our text classification experiment, we use a dataset consisting of hand-selected paragraphs from companies' annual reports or sustainability reports. All paragraphs were annotated as \emph{yes} (climate-related) or \emph{no} (not climate-related) by at least four experts from the field using the software prodigy.\footnote{\url{www.prodi.gy}} See Appendix \ref{app:annotation_guidelines_climate} for our annotation guidelines. In case of a close verdict or a tie between the annotators, the authors of this paper discussed the paragraph in depth before reaching an agreement.

In the following, Table \ref{tbl:text_classification_results} reports the results of the language models when trained on our text classification task, i.e., whether the text is climate-related or not. Overall, we find that all \climatebert LMs outperform a non-pre-trained Distil\roberta across both metrics for the text classification task. Most notably, domain-adaptive pretraining with similar samples to our downstream tasks (${\climatebert}_S$) leads to improvements of 32.64\% in terms of cross-entropy loss and a reduction in the error rate of the F1 score by 35.71\%.

\begin{table}[h]
\centering
\begin{tabular}{l cc}
		\toprule
		     & \multicolumn{2}{c}{\textbf{Text classification}} \\
		     \cmidrule{2-3}
		    {\textbf{Model}} & {\textbf{Loss}} & {\textbf{F1}} \\
		\midrule
		{Distil\roberta} & $0.242_{0.171}$ & $0.986_{0.010}$ \\[3pt]
        ${\climatebert}_F$ & $0.191_{0.136}$ & $0.989_{0.010}$ \\
        ${\climatebert}_S$ & $0.163_{0.132}$ & $0.991_{0.008}$ \\
        ${\climatebert}_D$ & $0.197_{0.153}$ & $0.988_{0.009}$ \\
        ${\climatebert}_{D + S}$ & $0.217_{0.153}$ & $0.988_{0.009}$ \\
        \bottomrule
\end{tabular}
\caption{Results on our text classification task. Reported are the average cross-entropy loss and the average weighted F1 score on the validation sets across 60 evaluation runs. Value subscripts report the standard deviations.}
\label{tbl:text_classification_results}
\end{table}

\subsection{Sentiment Analysis}

Our next task studies the sentiment behind the climate-related paragraphs, using the same dataset as in the previous section. In our context, we use the term `sentiment' to distinguish whether an entity reports on climate-related developments as negative \emph{risk}, as positive \emph{opportunity}, or as \emph{neutral}. 


Therefore, we created a second labeled dataset on climate-related sentiment, for which we asked the annotators to label the paragraphs by one of the three categories --- \emph{risk}, \emph{neutral}, or \emph{opportunity}. See Appendix \ref{app:annotation_guidelines_sentiment} for our annotation guidelines. Similarly, as before, in case of a close verdict or a tie between the annotators, the authors of this paper discussed the paragraph in depth before reaching an agreement.

Table \ref{tbl:sentiment_results} shows the performance of our models in sentiment prediction. Again, all \climatebert LMs outperform the Distil\roberta baseline model in terms of F1 score and average cross-entropy loss. The largest improvements can be observed with ${\climatebert}_F$, which amount to a 7.33\% lower cross-entropy loss and a 7.42\% lower error rate in terms of average F1 score compared to the Distil\roberta baseline LM. 

\begin{table}[h]
\centering
\begin{tabular}{l cc}
		\toprule
		     & \multicolumn{2}{c}{\textbf{Sentiment analysis}} \\
		     \cmidrule{2-3}
		    {\textbf{Model}} & {\textbf{Loss}} & {\textbf{F1}} \\
		\midrule
		{Distil\roberta} & $0.150_{0.069}$ & $0.825_{0.046}$ \\[3pt] 
        ${\climatebert}_F$ & $0.139_{0.042}$ & $0.838_{0.036}$ \\
        ${\climatebert}_S$ & $0.140_{0.057}$ & $0.836_{0.033}$ \\
        ${\climatebert}_D$  & $0.138_{0.043}$ & $0.835_{0.040}$ \\
        ${\climatebert}_{D + S}$ & $0.139_{0.043}$ & $0.834_{0.036}$ \\
        \bottomrule
\end{tabular}
\caption{Results on our sentiment analysis task in terms of average validation loss and average weighted F1 score across 60 evaluation runs. Subscripts report the standard deviations.}
\label{tbl:sentiment_results}
\end{table}

\subsection{Fact-Checking}

We now turn to the fact-checking downstream task. We apply our model to a dataset that was proposed by \citet{diggelmann2020climate} and comprises 1.5k sentences that make a claim about climate-related topics. This \climatefever dataset is to the best of our knowledge to date the only dataset that focuses on climate change fact-checking. \climatefever adapts the methodology of \fever, the largest dataset of artificially designed claims, to real-life claims on climate change collected online. The authors of  \climatefever find that the surprising, subtle complexity of modeling real-world climate-related claims provides a valuable challenge for general natural language understanding. Working with this dataset, \citet{chillrud2021evidence} recently introduced a novel semi-supervised training method to achieve a state-of-the-art (SotA) F1 score of 0.7182 on the fact-checking dataset \climatefever.

\begin{table}[h] 
\centering
\begin{tabular}{lp{5cm}}
		\toprule
		\textbf{Claim}: &  97\% consensus on human-caused global warming has been disproven. \\ 
        \vtop{\hbox{\strut \textbf{Evidence}}\hbox{\strut REFUTE}}: &  In a 2019 CBS poll, 64\% of the US population said that climate change is a ""crisis"" or a ""serious problem"", with 44\% saying human activity was a significant contributor.  \\ \midrule
        \textbf{Claim}: & 
        The melting Greenland ice sheet is already a major contributor to rising sea level and if it was eventually lost entirely, the oceans would rise by six metres around the world, flooding many of the world’s largest cities.\\
             \vtop{\hbox{\strut \textbf{Evidence}}\hbox{\strut SUPPORT}}: &  The Greenland ice sheet occupies about 82\% of the surface of Greenland, and if melted would cause sea levels to rise by 7.2 metres.
        \\ \bottomrule
\end{tabular}
\caption{Examples taken from \climatefever.}
\label{tbl:climatefeverex}
\end{table}

Each claim in \climatefever is supported or refuted by evidence sentences (see Table \ref{tbl:climatefeverex}), and an evidence sentence can also be classified as giving not enough information. The objective of the model is to classify an evidence sentence to \emph{support} or \emph{refute} a claim. To feed this combination of claim and evidence into the model, we concatenate the claims with the related evidence sentences, with a $\textsc{[SEP]}$ token separating them. As in \citet{chillrud2021evidence}, and for comparison with their results, we filter out all evidence sentences with the label \emph{NOT\_ENOUGH\_INFO} in the \climatefever dataset.

Table \ref{tbl:climatefever} lists the results of our experiments on the \climatefever dataset. In line with our previous experiments, we find similar or better results for all \climatebert LMs across all metrics. Our ${\climatebert}_{D + S}$ LM achieves similar cross-entropy loss compared to the basis Distil\roberta model, yet pushes the average F1 score from 0.748 to 0.757, which outperforms \citet{chillrud2021evidence}'s previous SotA F1 score of 0.7182, and is hence, to the best of our knowledge, the new SotA on this dataset.

\begin{table}[h]
\centering
\begin{tabular}{l cc}
		\toprule
		     & \multicolumn{2}{c}{\textbf{Fact-checking}} \\
		     \cmidrule{2-3}
		    {\textbf{Model}} & {\textbf{Loss}} & {\textbf{F1}} \\
		\midrule
		{Distil\roberta} & $0.135_{0.017}$ & $0.748_{0.036}$ \\[3pt]
        ${\climatebert}_F$ & $0.134_{0.020}$ & $0.755_{0.037}$ \\
        ${\climatebert}_S$ & $0.133_{0.017}$ & $0.753_{0.042}$ \\
        ${\climatebert}_D$ & $0.135_{0.016}$ & $0.752_{0.042}$ \\
        ${\climatebert}_{D + S}$ & $0.135_{0.018}$ & $0.757_{0.044}$ \\
        \bottomrule
\end{tabular}
\caption{Results on our fact-checking task on \climatefever in terms of average validation loss and average weighted F1 score across 60 evaluation runs. Subscripts report the standard deviations.}
\label{tbl:climatefever}
\end{table}

\section{Carbon Footprint}
Training deep neural networks in general and large language models in particular, has a significant carbon footprint already today. If the LM research trends continue, this detrimental climate impact will increase considerably. The topic of efficient NLP was also discussed by a working group appointed by the ACL Executive Committee to promote ways that the ACL community can reduce the computational costs of model training (\url{https://public.ukp.informatik.tu-darmstadt.de/enlp/Efficient-NLP-policy-document.pdf}). We acknowledge that our work is part of this trend. In total, training \climatebert caused 115.15 kg CO2 emissions. We use two energy efficient NVIDIA RTX A5000 GPUs: 0.7 kW (power consumption of GPU server) x 350 hours (combined training time of all experiments) x 470 gCO2e/kWh (emission factor in Germany in 2018 according to \url{www.umweltbundesamt.de/publikationen/entwicklung-der-spezifischen-kohlendioxid-7}) = 115,149 gCO2e. We list all details about our climate impact in Table~\ref{tbl:model_card} in Appendix~\ref{app:model_card}.
Nevertheless, we decided to carry out this project, as we see the high potential of NLP to support action against climate change. Given our awareness of the carbon footprint of our research, we address this sensitive topic as follows: 
\begin{enumerate}
    \item We specifically decided to focus on Distil\roberta, which is a considerably smaller model in terms of number of parameters compared to the non-distilled version and, thus, requires less energy to train. Moreover, we do not crawl huge amounts of data without considering the quality. This way, we try to take into account the issues mentioned by \citet{bender2021dangers}.
    \item Hyperparameter tuning yields considerably higher CO2 emissions in the training stage due to tens or hundreds of different training runs. Note that our multiple training runs on the downstream task are not causing long training times as the downstream datasets are very small compared to the dataset used for training the language model. We therefore refrain from exhaustive hyperparameter tuning. Rather, we build on previous findings. We systematically experimented with a few hyperparameter combinations and found that the hyperparameters proposed by \citet{gururangan2020don} lead to the best results.
    \item We would have liked to train and run our model on servers powered by renewable energy. This first best option was unfortunately not available. In order to speed up the energy system transformation required to achieve the global climate targets, we contribute our part by donating Euro 100 to atmosfair. atmosfair was founded in 2005 and is supported by the German Federal Environment Agency. atmosfair offsets carbon dioxide in more than 20 locations: from efficient cookstoves in Nigeria, Ethiopia and India to biogas plants in Nepal and Thailand to solar energy in Senegal and Brazil and renewable energies in Tansania and Indonesia. See \url{www.atmosfair.de/en/offset/fix/}. We explicitly refrain from calling this donation a CO2 compensation, and we refrain from a solution that is based on afforestation.
\end{enumerate}

\section{Conclusion}
We propose \climatebert, the first language model that was pretrained on a large scale dataset of over 2 million climate-related paragraphs. We study various selection strategies to find samples from our corpus which are most helpful for later tasks. Our experiments reveal that our domain-adaptive pretraining leads to considerably lower masked language modeling loss on our climate corpus. We further find that this improvement is also reflected in predictive performance across three essential downstream climate-related NLP tasks: text classification, the analysis of risk and opportunity statements by corporations, and fact-checking climate-related claims.

\section*{Acknowledgments}
We are very thankful to Jan Minx and Max Callaghan from the Mercator Research Institute on Global Commons and Climate Change (MCC) Berlin for providing us with the data, which is a subset of the data they used in \citet{berrang2021systematic} and \citet{callaghan2021machine}.

\bibliography{acl2020}

\appendix

\section*{Appendix}

\section{Climate Performance Model Card}
\label{app:model_card}
Table~\ref{tbl:model_card} shows our climate performance model card, following \citet{hershcovich2022towards}.

\begin{table}[h]
\small
\begin{tabular}{@{}p{55mm}p{25mm}@{}}
\toprule
\multicolumn{2}{c}{\textbf{ClimateBert}} \\
\midrule
1. Model publicly available? & Yes \\
2. Time to train final model &  48 hours \\
3. Time for all experiments &  350 hours \\
4. Power of GPU and CPU & 0.7 kW \\
5. Location for computations & Germany \\
6. Energy mix at location & 470 gCO2eq/kWh \\
7. CO2eq for final model & 15.79 kg \\
8. CO2eq for all experiments & 115.15 kg \\
9. Average CO2eq for inference per sample & 0.62 mg \\
\bottomrule
\end{tabular}
\caption{Climate performance model card for ClimateBert.}
\label{tbl:model_card}
\end{table}

\section{Annotation Guidelines}

For our annotation procedure, we implemented the following general rules. The annotators had to label climate-relevant paragraphs. If the paragraph was climate-relevant, then they had to attach a sentiment to a paragraph. Annotators were asked to apply common sense, e.g., when a given paragraph might not provide all the context, but the context might seem obvious. Moreover, annotators were informed that each annotation should be a 0-1 decision. Hence, if an annotator was 70\% certain, then this was rounded up to 100\%. We asked, on average, five researchers to annotate the same tasks to obtain some measure of dispersion.  In case of a close verdict or a tie between the annotators, the authors of this paper discussed the paragraph in depth before reaching an agreement.

\subsection{Text classification}
\label{app:annotation_guidelines_climate}

The first task was to label climate-relevant paragraphs. The labels are \emph{Yes} or \emph{No}. As a general rule, we determined that just discussing nature/environment can be sufficient, and mentioning clean energy, emissions, fossil fuels, etc., can also be sufficient. It is a \emph{Yes}, if the paragraph includes some wording on a climate change or environment related topic (including transition and litigation risks, i.e., emission mitigation measures, energy consumption and energy sources etc.; and physical risks, i.e., increase in risk of floods, coastal area exposure, storms etc.). It is a \emph{No}, if the paragraph is not related to climate policy, climate change or an environmental topic at all. For some examples, see Table \ref{tbl:annotation_climate}.

\begin{table}[t!] 
\centering
\begin{tabular}{lp{5cm}}
		\toprule
		     {\textbf{Label}} &  {\textbf{Examples}} \\
		\midrule
		Yes &  
        Sustainability: The Group is subject to stringent and evolving laws, regulations, standards and best practices in the area of sustainability (comprising corporate governance, environmental management and climate change (specifically capping of emissions), health and safety management and social performance) which may give rise to increased ongoing remediation and/or other compliance costs and may adversely affect the Group’s business, results of operations, financial condition and/or prospects. \\ \midrule
        Yes & Scope 3: Optional scope that includes indirect emissions associated with the goods and services supply chain produced outside the organization. Included are emissions from the transport of products from our logistics centres to stores (downstream) performed by external logistics operators (air, land and sea transport) as well as the emissions associated with electricity consumption in franchise stores.\\ \midrule
        No & 
        Risk and risk management Operational risk and compliance risk Operational risk is the risk of loss resulting from inadequate or failed internal processes, people and systems, or from external events including legal risk but excluding strategic and reputation risk. It also includes, among other things, technology risk, model risk and outsourcing risk.
        \\ \bottomrule
\end{tabular}
\caption{Examples for the annotation task climate (Yes/No).}
\label{tbl:annotation_climate}
\end{table}

\subsection{Sentiment Analysis}
\label{app:annotation_guidelines_sentiment}

For the sentiment analysis, annotators had to provide labels as to whether a (climate change-related) paragraph talks about a \emph{Risk} or threat that negatively impacts an entity of interest, i.e. a company (negative sentiment), or whether an entity is referring to some \emph{Opportunity} arising due to climate change (positive sentiment). The paragraph can also make just a \emph{Neutral} statement. 

To be more precise, we consider a paragraph relating to risk, if the paragraph mainly talks about 1) business downside risks, potential losses and adverse developments detrimental to the entity 2) and/or about negative impact of an entity's activities on the society/environment 3) and/or associates specific negative adjectives to the anticipated, past or present developments and topics covered. 

We consider a paragraph relating to opportunities, if the paragraph mainly talks about 1) business opportunities arising from mitigating climate change, from adapting to climate change etc. which might be beneficial for a specific entity 2) and/or about positive impact of an entity's activities on the society/environment 3) and/or associates specific positive adjectives to the anticipated, past or present developments and topics covered. 

Lastly, we consider a paragraph as neutral if it mainly states facts and developments 1) without putting them into positive or negative perspective for a specific entity and/or the society and/or the environment, 2) and/or does not associate specific positive or negative adjectives to the anticipated, past or present facts stated and topics covered. For some examples, see Table \ref{tbl:annotation_sentiment}.
 
\begin{table}[t!]
\scriptsize
\centering
\begin{tabular}{lp{5cm}}
		\toprule
		     {\textbf{Label}} &  {\textbf{Examples}} \\
		\midrule
		Opportunity &   
		Grid \& Infrastructure and Retail – today represent the energy world of tomorrow. We rank among Europe‘s market leaders in the grid and retail business and have leading positions in renewables. We intend to spend a total of between Euro 6.5 billion and Euro 7.0 billion in capital throughout the Group from 2017 to 2019.  \\ \midrule 
		
		Opportunity &   
		We want to contribute to the transition to a circular economy. The linear economy is not sustainable. We discard a great deal (waste and therefore raw materials, experience, social capital and knowledge) and are squandering value as a result. This is not tenable from an economic and ecological perspective. As investor we can ‘direct’ companies and with our network, our scale and our influence we can help the movement towards a circular future (creating a sustainable society) further along. \\ \midrule 

        Neutral &  A similar approach could be used for allocating emissions in the fossil fuel electricity supply chain between coal miners, transporters and generators. We don’t invest in fossil fuel companies, but those investors who do should account properly for their role in the production of dangerous emissions from burning fossil fuels. \\ \midrule 
        
        Neutral & Omissions: Emissions associated with joint ventures and investments are not included in the emissions disclosure as they fall outside the scope of our operational boundary. We do not have any emissions associated with heat, steam or cooling. We are not aware of any other material sources of omissions from our emissions reporting.\\ \midrule 
        
        Risk &   We estimated that between 36.5 and 52.9 per cent of loans granted to our clients are exposed to transition risks. If the regulator decides to pass ambitious laws to accelerate the transition towards a low-carbon economy, carbon-intensive companies would incur in higher costs, which may prevent them from repaying their debt. In turn, this would weaken our bank’s balance sheets. . \\ \midrule 
        
        Risk & American National Insurance Company recognizes that increased claims activity resulting from catastrophic events, whether natural or man-made, may result in significant losses, and that climate change may also affect the affordability and availability of property and casualty insurance and the pricing for such products.         \\
        \bottomrule
\end{tabular}
\caption{Examples for the annotation task sentiment (Opportunity/Neutral/Risk).}
\label{tbl:annotation_sentiment}
\end{table}  

\section{Added Tokens}
\label{sec:added_tokens}

'CO2', 'emissions', '’', 'temperature', 'environmental', 'soil', 'increase', 'conditions', 'potential', 'increased', 'areas', 'degrees', 'across', 'systems', 'emission', 'precipitation', 'impacts', 'compared', 'countries', 'sustainable', 'provide', 'reduction', 'annual', 'reduce', 'greenhouse', 'approach', 'processes', 'factors', 'observed', 'renewable', 'temperatures', 'distribution', 'studies', 'variability', 'significantly', '–', 'further', 'regions', 'addition', 'showed', '“', 'industry', 'consumption', 'regional', 'risks', 'atmospheric', 'supply', 'companies', 'plants', 'biomass', 'electricity', 'respectively', 'activities', 'communities', 'climatic', 'solar', 'investment', 'spatial', 'rainfall', '•', 'sustainability', 'costs', 'reduced', '2021', 'influence', 'vegetation', 'sources', 'possible', 'ecosystem', 'scenarios', 'summer', 'drought', 'structure', 'economy', 'considered', 'various', 'atmosphere', 'several', 'technologies', 'transition', 'assessment', 'dioxide', 'ocean', 'fossil', 'patterns', 'waste', 'solutions', 'transport', 'strategy', 'CH4', 'policies', 'understanding', 'concentration', 'customers', 'methane', 'applied', 'increases', 'estimated', 'flood', 'measured', 'thermal', 'concentrations', 'decrease', 'greater', 'following', 'proposed', 'trends', 'basis', 'provides', 'operations', 'differences', 'hydrogen', 'adaptation', 'methods', 'capture', 'variation', 'reducing', 'N2O', 'parameters', 'ecosystems', 'investigated', 'yield', 'strategies', 'indicate', 'caused', 'dynamics', 'obtained', 'efforts', 'coastal', 'become', 'agricultural', 'decreased', 'GHG', 'materials', 'mainly', 'relationship', 'ecological', 'benefits', '+/-', 'challenges', 'nitrogen', 'forests', 'trend', 'estimates', 'towards', 'Committee', 'seasonal', 'developing', 'particular', 'importance', 'tropical', 'ratio', '2030', 'composition', 'employees', 'characteristics', 'scenario', 'measurements', 'plans', 'fuels', 'infrastructure', 'overall', 'responses', 'presented', 'least', 'assess', 'diversity', 'periods', 'delta', 'included', 'already', 'targets', 'achieve', 'affect', 'conducted', 'operating', 'populations', 'variations', 'studied', 'additional', 'construction', 'northern', 'variables', 'soils', 'ensure', 'recovery', 'combined', 'decision', 'practices', 'however', 'determined', 'resulting', 'mitigation', 'conservation', 'estimate', 'identify', 'observations', 'losses', 'productivity', 'agreement', 'monitoring', 'investments', 'pollution', 'contribution', 'opportunities', 'simulations', 'gases', 'statements', 'planning', 'shares', 'sediment', 'flux', 'requirements', 'trees', 'temporal', 'determine', 'southern', 'previous', 'integrated', 'relatively', 'analyses', 'means', '2050', '”', 'uncertainty', 'pandemic', 'fluxes', 'findings', 'moisture', 'consistent', 'decades', 'snow', 'performed', 'contribute', 'crisis'

\end{document}